\documentclass[10pt,twocolumn,letterpaper]{article}

\usepackage{iccv}
\usepackage{times}
\usepackage{epsfig}
\usepackage{graphicx}
\usepackage{amsmath}
\usepackage{amssymb}

\usepackage{booktabs}

\usepackage{bbm}
\usepackage{bm}

\usepackage{multirow}
\usepackage{subfig}

\def\ie{\mbox{\textit{i.e.}}}
\def\eg{\mbox{\textit{e.g.}}}
\def\wrt{\mbox{\textit{w.r.t.~}}}
\def\etc{\mbox{\textit{etc}}}

\newcommand{\qi}{\textcolor{black}}

\usepackage[pagebackref=true,breaklinks=true,letterpaper=true,colorlinks,bookmarks=false]{hyperref}

\iccvfinalcopy 

\def\iccvPaperID{6590} 

\ificcvfinal\fi

\begin{document}

\title{Likelihood-Based Text-to-Image Evaluation with\\Patch-Level Perceptual and Semantic Credit Assignment}

\author{Qi Chen$^{1,*}$, Chaorui Deng$^{1,*}$, Zixiong Huang$^2$, Bowen Zhang$^1$, Mingkui Tan$^2$, Qi Wu$^{1,\dagger}$\\
$^1$Australia Institute of Machine Learning, University of Adelaide\\
$^2$South China University of Technology\\
}

\maketitle
\ificcvfinal\thispagestyle{empty}\fi

\makeatletter
\def\blfootnote{\xdef\@thefnmark{}\@footnotetext}
\makeatother

\begin{abstract}

Text-to-image synthesis has made encouraging progress and attracted lots of public attention recently. However, popular evaluation metrics in this area, like the Inception Score and Fr\'{e}chet Inception Distance, incur several issues.
First of all, they cannot explicitly assess the perceptual quality of generated images and poorly reflect the semantic alignment of each text-image pair.
Also, they are inefficient and need to sample thousands of images to stabilise their evaluation results.
In this paper, we propose to evaluate text-to-image generation performance by directly estimating the likelihood of the generated images using a pre-trained likelihood-based text-to-image generative model,
\ie, a higher likelihood indicates better perceptual quality and better text-image alignment.
To prevent the likelihood of being dominated by the non-crucial part of the generated image, 
we propose several new designs to develop a credit assignment strategy based on the semantic and perceptual significance of the image patches.
In the experiments, we evaluate the proposed metric on multiple popular text-to-image generation models and datasets in accessing both the perceptual quality and the text-image alignment.
Moreover, it can successfully assess the generation ability of these models with as few as a hundred samples,
making it very efficient in practice.
\blfootnote{$^*$Co-first author. $^\dagger$Corresponds to~\texttt{qi.wu01@adelaide.edu.au}. Code is available at \texttt{https://github.com/chenqi008/LEICA}.}
\end{abstract}

\section{Introduction}\label{sec:intro}
Recent works \cite{ding2022cogview2,esser2021taming,nichol2021glide,ramesh2021zero,ramesh2022hierarchical,saharia2022photorealistic,yu2022scaling} 
have made promising achievements on the text-to-image generation task, 
which can produce photo-realistic images or impressive artworks according to the text inputs.
However, automatic evaluation of the text-to-image generation performance has always been a challenging problem, especially in complex, open-vocabulary scenarios.
Most existing text-to-image generation models, including those state-of-the-art models, still rely on the popular evaluation metrics in general image generation,
such as the Inception Score (IS) \cite{salimans2016improved} and Fr\'{e}chet Inception Distance (FID) \cite{heusel2017gans}.

The IS leverages a pre-trained image recognition model (Inception-V3 \cite{szegedy2016rethinking}) 
to provide the recognition confidence of the generated images to assess the generative performance.
A higher confidence score may indicate that the generated images are more similar to those used for training the Inception-V3 (\ie, ImageNet~\cite{deng2009imagenet}). 
However, the critical issue is that the similarity with the pre-training images does not necessarily reflect the perceptual quality or semantic alignment of the generated images,
since there exists a significant domain shift between the pre-training data (which has relatively limited semantic diversity) and the open-vocabulary natural languages.
Besides, the pre-trained model can also be over/under-fitted. Thus, the confidence scores may be unreliable.
Likewise, the followers of IS, such as Classification Accuracy Score (CAS) \cite{ravuri2019classification} and Semantic Object Accuracy (SOA) \cite{hinz2020semantic}, could also suffer from these problems.

The FID, on the other hand, computes the distribution distance between the real images and the generated images in a latent space,
where the backbone of the Inception-V3 is used to map those images into the latent space, 
and a multivariate Gaussian assumption is imposed on the latent space to estimate the distribution statistics.
Although it addresses the domain shift problem of IS (to some extent) by considering real images during evaluation, FID still faces several issues.
First, the assumption of the multivariate Gaussian distribution is groundless, making it a poor indicator of the perceptual quality of generated images.
Second, FID is designed only to consider the overall similarity of two sets of images, 
which cannot measure the semantic alignment between each input text and the corresponding generated image.
Finally, both IS and FID is inefficient, as they have to process lots of image samples using the Inception V3 model to stabilize their evaluation results.
Most of these drawbacks can also be found in Kernel Inception Distance (KID) \cite{binkowski2018demystifying}, a kernelized variant of FID.

Tackle these problems are challenging. 
In this paper, we propose a new metric for text-to-image evaluation based on likelihood estimation.
Specifically, given $n$ image descriptions $X=\{x_i\}_{i=1}^n$ and the corresponding generated images $Y=\{y_i\}_{i=1}^n$,
we leverage a pre-trained likelihood-based text-to-image model $\mathcal{G}$ to estimate the likelihood of each $y_i$ given $x_i$ as condition, \ie, $P_{\mathcal{G}}(y_i|x_i)$.
\textcolor{black}{Ideally, when the model-estimated density is closely aligned with the real-world data density, 
then $P_{\mathcal{G}}(y_i|x_i)$ would effectively indicate how likely the image $y_i$ exists in real-world when the text $x_i$ is given~\cite{theis2015note}.
In this sense, to achieve a high $P_{\mathcal{G}}(y_i|x_i)$, $y_i$ should not only be semantically aligned with $x_i$, but also be perceptually good enough as the real-world images.
Otherwise, we will get a low likelihood score for $y_i$.
In fact, average log-likelihood has been the \textit{de facto} standard and widely considered as the default measure for detecting out-of-distribution data~\cite{ren2019likelihood} or quantifying generative modeling performance \cite{theis2015note,van2016pixel}. 
However, existing text-to-image evaluation methods have to resort to the aforementioned alternatives due to the lack of a well-trained likelihood estimator $\mathcal{G}$.}

We address this problem using the recently emerged likelihood-based autoregressive image generation models \cite{ramesh2021zero,wang2022ofa}, 
which have shown impressive text-to-image generation performance. 
Concretely, they first rely on a discrete variational autoencoder (dVAE) \cite{van2017neural,esser2021taming} to learn a codebook so that an image can be converted into a sequence of visual codes and vice versa.
The generative modeling is then accomplished by a transformer model \cite{vaswani2017attention},
which takes the text $x_i$ as input and predicts the likelihood of the visual codes autoregressively for the image $y_i$.
In this way, we can obtain $P_\mathcal{G}(y_i|x_i)$ as the product of the likelihood of the visual codes.
In the evaluation setting, $y_i$ is given, and its visual codes can be directly obtained from the dVAE. Thus, their likelihoods can be efficiently predicted in parallel with just one forward pass of the transformer, \textcolor{black}{instead of the step-by-step autoregressive manner.}

Nevertheless, we observe that even with such a good estimator, $P_\mathcal{G}(y_i|x_i)$ can still be dominated by: 
1) The visual codes whose corresponding patterns occasionally appear in the generated images but rarely exist in real-world scenes.
We refer to them as visual codes with \textbf{low perceptual significance}.
Although they may not affect the overall perceptual quality of $y_i$ as they generally account for only a small ratio in the whole image,
they tend to have extremely small likelihoods and drag down the overall likelihood $P_\mathcal{G}(y_i|x_i)$;
2) The visual codes that are semantically irrelevant to the text, \ie, those corresponding to the background.
which exhibits \textbf{low semantic significance}, but usually accounts for the largest portion of the image and thus affects $P_\mathcal{G}(y_i|x_i)$ mostly.

To tackle these issues, we explicitly introduce a credit assignment strategy based on the perceptual and semantic significance of each visual code.
Specifically, we first design an indicator function to eliminate the factors of the rare visual codes from $P_\mathcal{G}(y_i|x_i)$, 
so that only the perceptually significant codes are considered.
Then, we design a semantic scoring function using a pre-trained image-text matching model CLIP \cite{radford2021learning} to down-weight the visual codes with low semantic significance.
Since the original CLIP can only predict an alignment score for the whole image and the text, 
we make non-trivial modifications to make it suitable for patch-level image-text matching, including a new inference-time model architecture and a new scoring function.
We denote our proposed metric as \textbf{L}ikelihood-based t\textbf{E}xt-to-\textbf{I}mage evaluation with \textbf{C}redit \textbf{A}ssignment (LEICA).

We verify the effectiveness of the proposed LEICA score on \textbf{11} state-of-the-art text-to-image generation models, 
including three Generative Adversarial Network (GAN) based models~\cite{zhou2022towards,hinz2020semantic,ye2022recurrent}, four autoregressive models~\cite{ramesh2021zero,ramesh2022hierarchical,ding2021cogview,ding2022cogview2}, and four diffusion models~\cite{rombach2022high,gu2022vector,saharia2022photorealistic,nichol2021glide}.
\textcolor{black}{We also evaluate LEICA on multiple text-to-image datasets, including one open-domain dataset COCO~\cite{lin2014microsoft}, and two specific-domain datasets CUB~\cite{wah2011caltech} and Oxford-Flower~\cite{nilsback2008automated}.}
We provide extensive analysis, including human studies, to show the superiority of our evaluation metric.

\section{Related Works}

\paragraph{Text-to-Image Generation}
models can be roughly categorized into three types: Generative Adversarial Networks (GAN), likelihood-based models, and score-based models.
Early works in this area are primarily GAN-based, such as StackGAN \cite{zhang2017stackgan}, AttnGAN \cite{xu2018attngan}, DM-GAN \cite{zhu2019dm}, and OP-GAN \cite{hinz2020semantic}.
However, GAN is essentially to model the sampling process of a distribution, which means we cannot directly obtain the likelihoods of the generated samples or anything that is related to the generation performance, and thus cannot be used for text-to-image evaluation.
For score-based models, a representative line of works is the recently popular diffusion models \cite{rombach2022high,gu2022vector,saharia2022photorealistic,nichol2021glide}, which achieve excellent performance in generating photo-realistic images and artworks.
Although diffusion models are able to obtain the likelihood of the generated image, it requires multiple forward steps of the model using inputs with different noise levels, which cannot be parallelized when estimating the likelihood of a generated image.
The dominant likelihood-based models, such as DALL-E \cite{ramesh2021zero}, OFA \cite{wang2022ofa}, CogView \cite{ding2021cogview,ding2022cogview2}, and Parti \cite{yu2022scaling}, are all autoregressive models.
They not only perform as well as diffusion models, but more importantly, given a generated image, they can efficiently obtain its likelihood within one forward step, making them very suitable for evaluating text-to-image generation. 

\paragraph{Evaluation Metrics for Text-to-Image Generation.}
As we have discussed in Section \ref{sec:intro}, the widely used metrics for generative image modeling, such as IS \cite{salimans2016improved}, CAS \cite{ravuri2019classification}, FID \cite{heusel2017gans}, and KID \cite{binkowski2018demystifying}, tend to be problematic in text-to-image evaluation.
SOA \cite{hinz2020semantic} is specifically designed for text-to-image generation on the MSCOCO \cite{lin2014microsoft} dataset.
It first filters all captions in the validation set for the keywords that are related to the object categories in MSCOCO.
For each filtered caption, several images are generated and fed into an object detector (YOLOv3 \cite{redmon2018yolov3} trained on MSCOCO) to see if any keywords-related object can be detected.
The average recall of these keywords over all categories or images is reported as the SOA-C and SOA-I scores, respectively.
Although it can evaluate the semantic alignment between the captions and the generated images, it is inconvenient to use (due to caption pre-processing) and not generalizable (captions and detectors are all bound to MSCOCO). Besides, it has to face all the problems that IS encounters, as mentioned in Section \ref{sec:intro}.
Some retrieval metrics, like R-precision, are used only to evaluate the semantic alignment of the generated images \cite{xu2018attngan} while ignoring the perceptual quality.
\qi{In contrast, average log-likelihood avoids the above phenomena, which is also widely considered as the default measure for evaluating generative models~\cite{theis2015note,van2016pixel}.}

\section{Method}

\paragraph{Overview} The overall pipeline of our evaluation metric LEICA consists of a likelihood estimation step and two credit assignment steps on the perceptual and semantic significance, respectively.
First of all, the likelihood of the generated image $y_i$ is estimated by
\begin{equation}\label{eq:likelihood}
P_{\mathcal{G}}(y_i|x_i)=\prod_{t=1}^mP_{\mathcal{G}}(c_{i,t}|c_{i,<t},x_i),
\end{equation}
where $\{c_{i,t}\}_{t=1}^m$ are the visual codes obtained from the dVAE given $y_i$ as input. 
Each $c_{i,t}$ is mainly responsible for representing the information nearby a specific spatial (patch) location in $y_i$, and $t$ is the flattened index of the patch.
$m$ is the total number of patches,
and $c_{i,<t}$ indicates the codes before $t$, \ie, $c_{i,1},...,c_{i,t-1}$.
\textcolor{black}{Note that, Eq.~(\ref{eq:likelihood}) serves as the basis of the proposed evaluation metric, where this vanilla likelihood estimation can provide a generally unbiased but unstable evaluation due to the existence of perceptually \& semantically insignificant visual codes.}

Then, we introduce an indicator function $H$ to explicitly eliminate the visual codes with low perceptual significance from Eq. (\ref{eq:likelihood}), \ie,
\begin{equation}\label{eq:indicator1}
\frac{1}{m}\sum_{t=1}^mH(P_{\mathcal{G}}(c_{i,t}|c_{i,<t},x_i)).
\end{equation}
We convert the likelihoods into log-likelihoods inside $H$ so the summation sign is used here, and the scores are averaged among all codes.
Lastly, we leverage the CLIP \cite{radford2021learning}, a well-trained image-text matching model (denoted by $\mathcal{M}$) to focus on the locations with high semantic significance.
As the CLIP can only provide the global matching score between an image-text pair, 
we modify its inference-time architecture to get $\mathcal{M}^\prime$, and further devise a scoring function $S$,
so as to assign a semantic alignment score for each location $t$.
The final evaluation score of LEICA on a sample ($x_i$, $y_i$) is then computed by
\begin{equation}
\label{eq:LECICA}
\text{LEICA}(x_i,y_i) = \frac{1}{m}\sum_{t=1}^m S(x_i, y_i, t) H(P_{\mathcal{G}}(c_{i,t}|c_{i,<t},x_i)).
\end{equation}
More details are in the following.

\begin{figure}[t]
    \centering
    \subfloat[Stable Diffusion]{\includegraphics[width=0.9\linewidth]{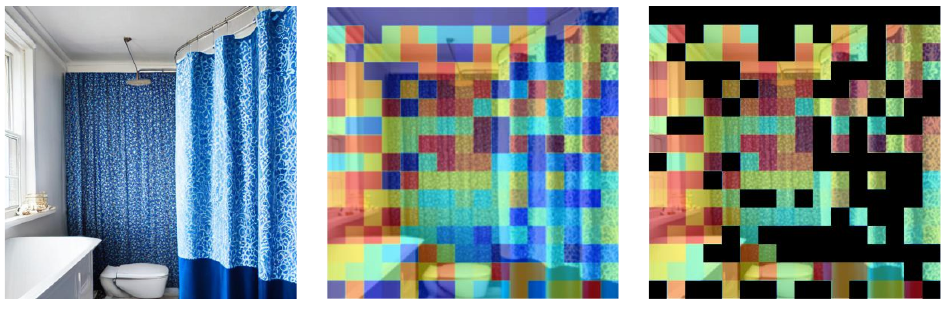}}\\
    \subfloat[OPGAN]{\includegraphics[width=0.9\linewidth]{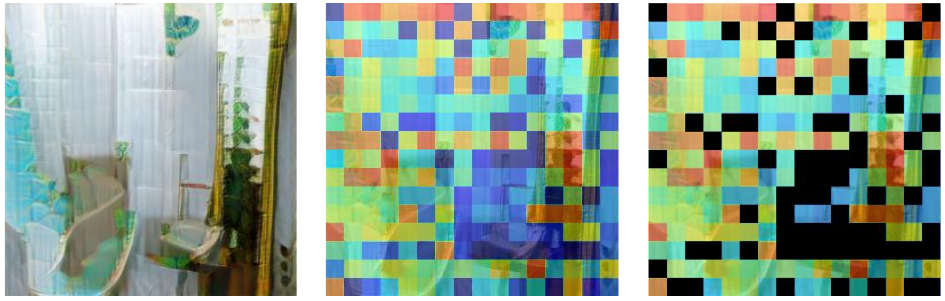}}
    \caption{Generated images for the caption \textit{``A bathroom with a blue shower curtain and walls''} with OPGAN and stable diffusion (the first column) and their visualized likelihood maps (the second column). 
    Although stable diffusion produces a much better result, its average likelihood is $1.401\times10^{-4}$, less than OPGAN ($8.020\times 10^{-4}$). 
    After removing the rare codes (the last column), their likelihoods increase to $3.297\times10^{-3}$ and $1.670\times10^{-3}$, respectively, where stable diffusion is considered to be better.} 
    \label{perceptual_significance}
\end{figure}

\subsection{Evaluation via Likelihood Estimation}
To directly estimate the likelihood of a generated image $y_i$ conditioned on the image description $x_i$, we leverage a well-trained likelihood-based text-to-image generator $\mathcal{G}$, which consists of a dVAE model and a transformer model.

Specifically, the dVAE is originally learned to convert an image into a sequence of visual codes and vice versa, so that the generation of images can be reformulated as the generation of their code sequences.
It is a convolutional encoder-decoder architecture accompanied by a learnable codebook that contains $K$ visual codes.
In our evaluation setting, we can directly use the trained dVAE to tokenize the generated images.
Suppose $y_i \in \mathbb{R}^{3\times H\times W}$, the encoder of dVAE first maps it into a feature map $\bm{f}_i \in \mathbb{R}^{d\times h\times w}$ where $h,w<H,W$.
Then, each spatial location in $\bm{f}_i$ will be represented by its nearest code retrieved from the codebook, and $y_i$ is thus represented by $m=h\times w$ visual codes, \ie, $\{c_{i,t}\}_{t=1}^m$.
The decoder of dVAE is used for reconstructing $y_i$ from $\{c_{i,t}\}_{t=1}^m$, which is not required in our setting.

The transformer is trained to model the joint distribution between image descriptions and the code sequences of the corresponding images.
The joint distribution is explicitly optimized to have a causal chain structure, where each visual code is predicted conditioning only on previous visual codes and the text.
Thus, during the generation procedure, the visual codes are predicted in an autoregressive manner.
When it comes to our evaluation setting, both $x_i$ and $\{c_{i,t}\}_{t=1}^m$ are given, we thus can obtain the conditional likelihood $P_\mathcal{G}(\{c_{i,t}\}_{t=1}^m|x_i)$ by one forward pass of the transformer, which serves as a lossless approximate to $P_\mathcal{G}(y_i|x_i)$.
Due to its causal chain structure, $P_\mathcal{G}(\{c_{i,t}\}_{t=1}^m|x_i)$ can be factorized as the product of the likelihood of each $c_{i,t}$, leading to the form in Eq. (\ref{eq:likelihood}).
We further consider using log-likelihood instead of likelihood to avoid too small values, \ie, $\ln P_{\mathcal{G}}(y_i|x_i)=\sum_{t=1}^m\ln P_{\mathcal{G}}(c_{i,t}|c_{i,<t},x_i)$.

\subsection{Credit Assignment of Perceptual Significance}
While it is feasible to directly assess the text-to-image generation performance using the likelihood estimated above,
we observed that there usually exists a small portion of codes in $\{c_{i,t}\}_{t=1}^m$ that have extremely small likelihoods (typically less than $10^{-9}$) 
and severely drag down the overall likelihood of $y_i$.
As shown in Figure \ref{perceptual_significance}, this phenomenon occurs no matter what the actual qualitative performance is.
As a result, a well-synthesized image may be rated lower than a poorly-synthesized image just because it contains more rarely-used visual codes.

To handle this problem, we introduce an indicator function $H$ on the likelihood of each visual code into Eq. (\ref{eq:likelihood}) to explicitly focus on those with high perceptual significance:
\begin{equation}\label{eq:indicator2}
\begin{split}
H(P_{\mathcal{G}}(c_{i,t}|c_{i,<t}, x_i)) = \mathbbm{1}(\ln P_{\text{data}}(c_{i,t}) - \lambda)& \\~*~ \max(\ln P_{\mathcal{G}}(c_{i,t}|c_{i,<t},x_i) - \lambda,0)&, 
\end{split}
\end{equation}
Here, $\mathbbm{1}(\cdot)$ equals 1 for positive inputs and 0 otherwise.
$P_\text{data}(c_{i,t})$ is the prior probability of $c_{i,t}$ estimated from real images.
$\lambda < 0$ is a threshold that controls whether the effect of $c_{i,t}$ should be considered or not according to both its prior probability $P_\text{data}(c_{i,t})$, \ie, the perceptual significance of $c_{i,t}$ in real images, and the predicted likelihood $P_{\mathcal{G}}(c_{i,t}|c_{i,<t},x_i)$, which indicates its perceptual significance in the generated image.
In this way, only the visual codes whose predicted likelihoods are larger than $e^\lambda$ are rewarded with positive values, and those values will only take effect when the visual codes have higher prior probability than $e^\lambda$. 
Otherwise, $H(P_{\mathcal{G}}(c_{i,t}|c_{i,<t}, x_i))$ will be zeroed out from the summation in Eq. (\ref{eq:indicator1}).
By default, $\lambda$ is set to a small value, \eg, $\ln 10^{-9}$. 
Thus, $H$ will only affect a few codes in $\{c_{i,t}\}_{t=1}^m$ that have extremely small likelihoods.

\subsection{Credit Assignment of Semantic Significance}

\begin{figure}[t]
    \centering
    \includegraphics[width=0.9\linewidth]{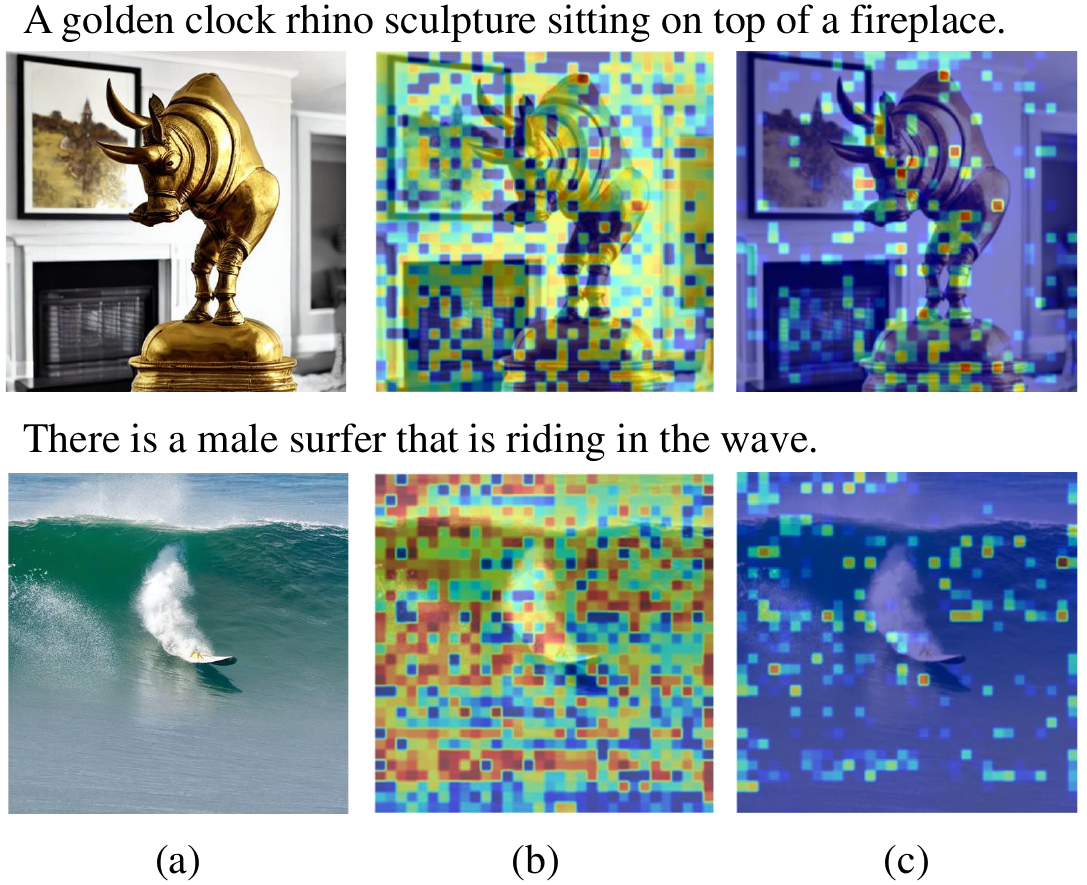}
    \caption{(a) are the images generated by the stable diffusion model. (b) and (c) are the likelihood maps before and after the semantic credit assignment, respectively. For these two images, their overall likelihoods are $2.422\times10^{-4}$ and $4.123\times10^{-4}$, respectively, where the second image which misses the foreground object \textit{``male surfer''} is rated higher than the first image.
    This can be corrected after the semantic credit assignment, where the overall likelihoods become $1.049\times10^{-4}$ and $4.704\times10^{-5}$ in (c).} 
    \label{semantic_significance}
\end{figure}

Since the background information (that is irrelevant to the input description $x_i$) usually constitutes a large part of the generated image $y_i$, it can easily dominate the overall likelihood, as shown in Figure \ref{semantic_significance}.
Therefore, it would be beneficial to focus on the semantically significant part of $y_i$, which is also very close to human behavior.
To achieve this, an intuitive idea is to use a pre-trained image-text matching model with zero-shot recognition abilities to estimate the semantic significance. 
We choose the CLIP~\cite{radford2021learning}, a multi-modality model trained on web-scale image-text pairs.

\paragraph{Adapt CLIP for Patch-Level Scoring}
The CLIP model we adopted (denoted by $\mathcal{M}$) has two transformer encoders to process images and texts, respectively. 
A feature vector is extracted from each modality with the corresponding encoder and is linearly projected into the same dimensional space and $L2$-normalized.
The optimization of $\mathcal{M}$ is achieved via contrastive learning, where the feature vectors that belong to the same image-text pair are pulled close, and the others are pushed away.
However, when used for our evaluation setting, although we can assess how the generated $y_i$ and the text $x_i$ are semantically close using the cosine similarity of their feature vectors, 
we are unfortunately unable to assess the alignment at a more detailed level, \ie, how each spatial location in $y_i$ aligns with $x_i$.
This is caused by the feature extraction mechanism of the image encoder. 
It takes $s\times s$ non-overlapping patches divided from the image and an extra \texttt{[CLS]} token as input, and the final embedding of the \texttt{[CLS]} token is adopted as the output image feature vector, while the final patch embeddings are all ignored.
Consequently, the contrastive loss is only applied on the \texttt{[CLS]} token, making the similarity between the text feature vector and the final patch embeddings uncertain during inference. 
We address this problem by proposing a new inference-time architecture for $\mathcal{M}$, termed as $\mathcal{M}^\prime$.

\begin{figure}[t]
    \centering
    \subfloat[CLIP]{\includegraphics[width=0.465\linewidth]{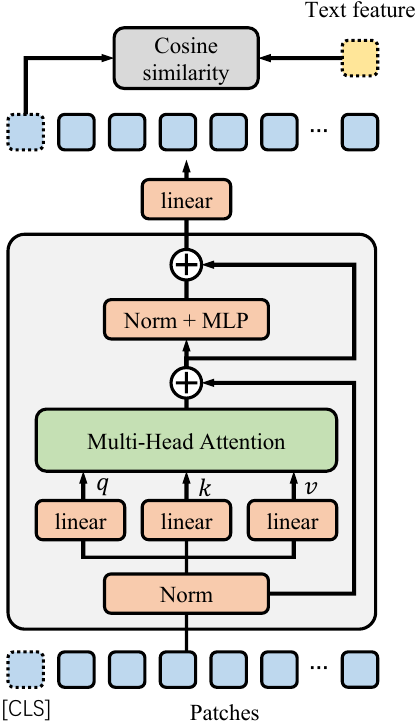}}~~~~~
    \subfloat[Ours]{\includegraphics[width=0.45\linewidth]{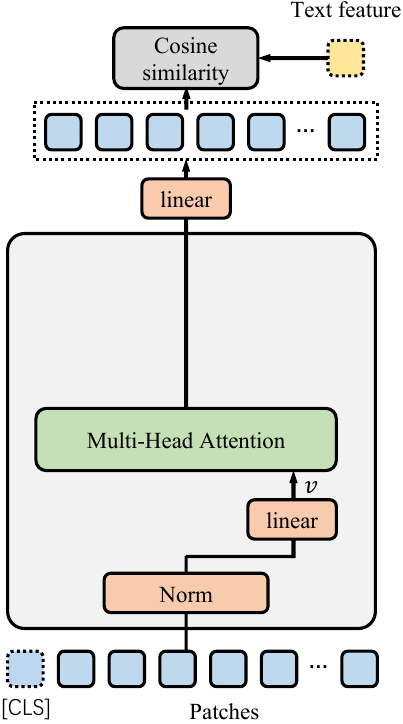}}
    \caption{The computation graph of the last layer in the original CLIP and our modified CLIP. } 
    \label{patch_level_arch}
\end{figure}

Our basic idea is to look for the patch representations in the computation graph of $\mathcal{M}$ that is closest to the loss computation node, which will also be the most relevant patch representations to the text feature vector.
As shown in Figure~\ref{patch_level_arch}, the last interaction between the patch embeddings and the \texttt{[CLS]} token embedding occurs within the last self-attention operation of the image encoder.
Thus, all the operations afterward, \eg, the last feed-forward network, the last layer normalization, \etc, are only used for the \texttt{[CLS]} token and have no benefit to the modeling of patch-level image-text relation.
We thus remove them from $\mathcal{M}$.
Moreover, inside the last self-attention operation, the \texttt{[CLS]} token is updated with the patch representations output by the value projection layer, denoted by $\{\bm{v}_{i,t}\}_{t=1}^{s^2}$, while the query and key projection layers are only used for computing the self-attention map.
Therefore, we leverage $\{\bm{v}_{i,t}\}_{t=1}^{s^2}$ to compute the patch-level alignment based on their cosine similarities with the feature vector of $x_i$, denoted by $\bm{x}_i$.
\begin{equation}
\phi_i(t) = \frac{(\bm{W}^\top\bm{v}_{i,t})^\top\bm{x}_{i}}{\|\bm{W}_1^\top\bm{v}_{i,t}\|_2^2\cdot\|\bm{x}_{i}\|_2^2}, ~ t= {1, ..., s^2}
\end{equation}
where $\bm{W}$ is the final linear project layer of the image encoder that transforms $\bm{v}_{i,t}$ into the same dimensional space as $\bm{x}_i$.
As shown in Figure \ref{fig:patch_mask}, with our modified CLIP model $\mathcal{M}^\prime$, the obtained patch-level similarity scores are closely aligned with the foreground part of the images.

\paragraph{Scoring Function for Semantic Significance}
To obtain the final semantic significance scores for the predicted visual codes $\{c_{i,t}\}_{t=1}^m$, we first apply a bilinear interpolation on $\phi_i$ to resize its sequence length from $s^2$ to $m$.
We then clip the negative values in $\phi_i$ to zeros so that the unrelated parts in $y_i$ will not affect the final evaluation score.
Finally, we can further consider the global image-text alignment score with the original CLIP model $\mathcal{M}$, to enhance the ability of our proposed LEICA metric on the semantic alignment evaluation.
Our scoring function for semantic significance is defined as follows:
\begin{equation}
S(x_i,y_i,t) = e^{\psi_i/\tau}\max(\phi_i(t), 0),
\label{eq:scoring_function}
\end{equation}
where $\psi_i$ is the cosine similarity between the feature vectors of $y_i$ and $x_i$. As the cosine similarities obtain from CLIP are generally within $(0.2, 0.4)$, we use the exponential function to scale the values to a broader range to distinguish the semantically-aligned/unaligned results better.
$\tau$ is a temperature hyper-parameter that controls the scaling effect, which is set to $0.07$ following \cite{chen2020simple}. The final LEICA score is calculated via Eq.~(\ref{eq:LECICA}).

\section{Experiments}

\paragraph{Meta Evaluation}
To assess the quality of the proposed evaluation metrics, following~\cite{yuan2021bartscore}, we use four measures for the meta-evaluation: 1) \textbf{Accuracy}, in our experiments, measures the percentage of correct judgment made by the metrics on whether a text is matched with an image or is mismatched/altered.
2) \textbf{Kendall's Tau Correlation} measures the ordinal association between two sets of data.
3) \textbf{Pearson Correlation} measures the linear correlation between two variables.
4) \textbf{Spearman Correlation} assesses the consistency between the ranking of two variables.

\paragraph{Implementation Details}

For the pre-trained likelihood-based text-to-image generation model $G$, 
we use OFA~\cite{wang2022ofa}, which is the best-performed and open-sourced autoregressive text-to-image generation model on MSCOCO~\cite{lin2014microsoft} dataset.
Still, we can always update it to the most powerful model in the future for better evaluation capability.
\qi{For the images used for evaluation, we generate them using the captions from the validation set of Karpathy’s split on MSCOCO, where the images and captions are semantically diverse enough and are also commonly used to evaluate the performance of text-to-image generation models in previous works~\cite{zhou2022towards,xu2018attngan,zhang2021cross,gu2022vector}.
Besides, we consider two datasets with specific domains: CUB~\cite{wah2011caltech} and Oxford-Flower~\cite{nilsback2008automated} to test the effectiveness of our score.}
We set the default resolution for the generated images to $256\times 256$, following the common practice in text-to-image generation.

\begin{figure}[t]
    \centering
    \subfloat[Gaussian Noise]{\includegraphics[width=0.5\linewidth]{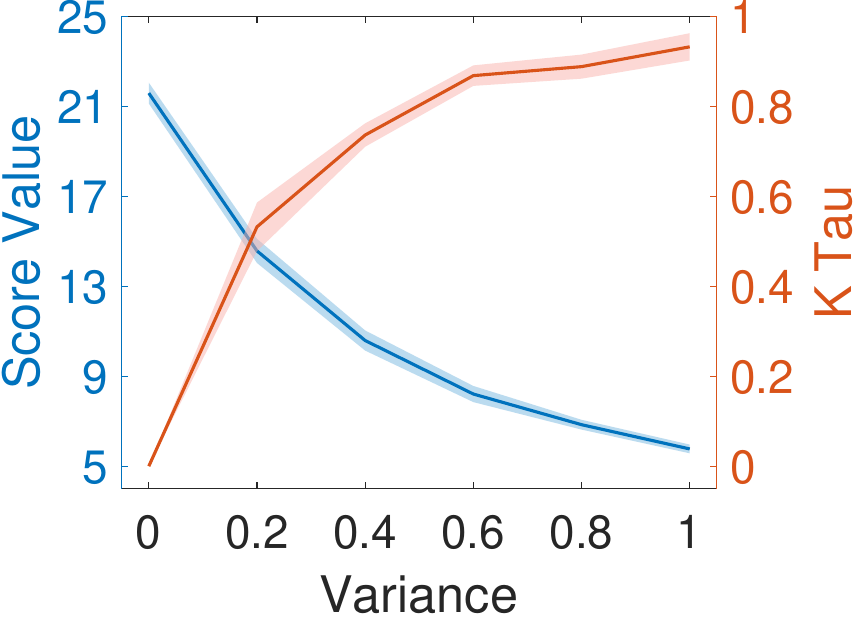}}~~
    \subfloat[Gaussian Blur]{\includegraphics[width=0.5\linewidth]{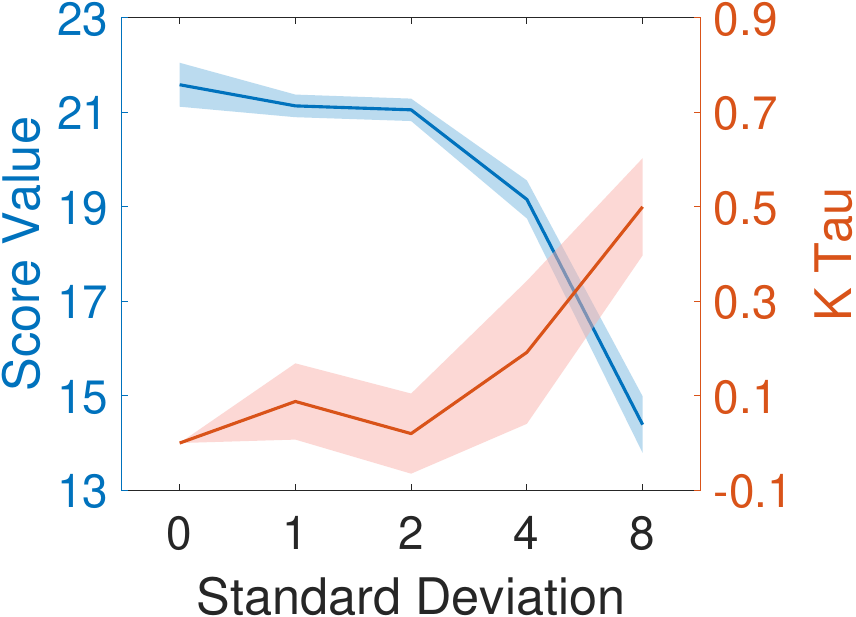}}~~\\
    \subfloat[Salt \& Pepper Noise]{\includegraphics[width=0.5\linewidth]{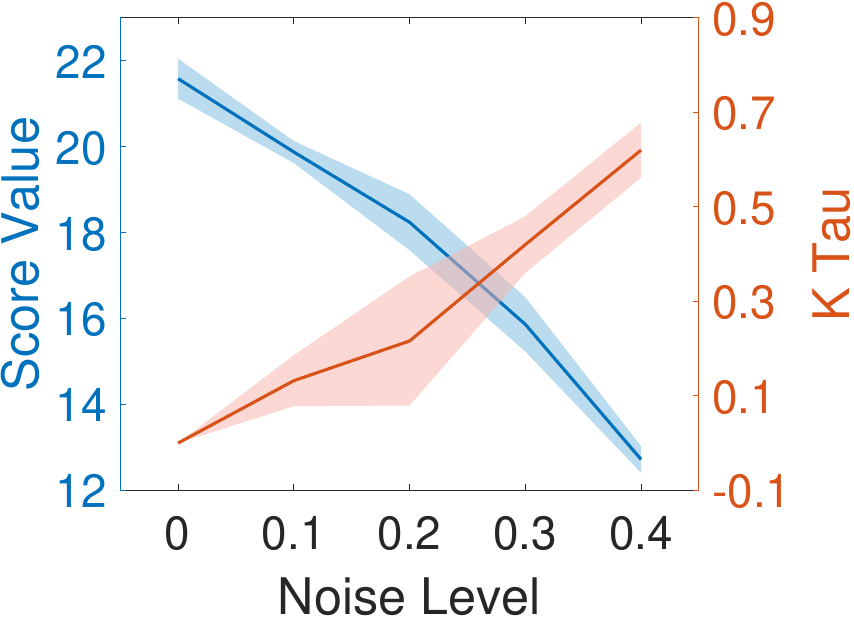}}~~
    \subfloat[Word Replacement]{\includegraphics[width=0.5\linewidth]{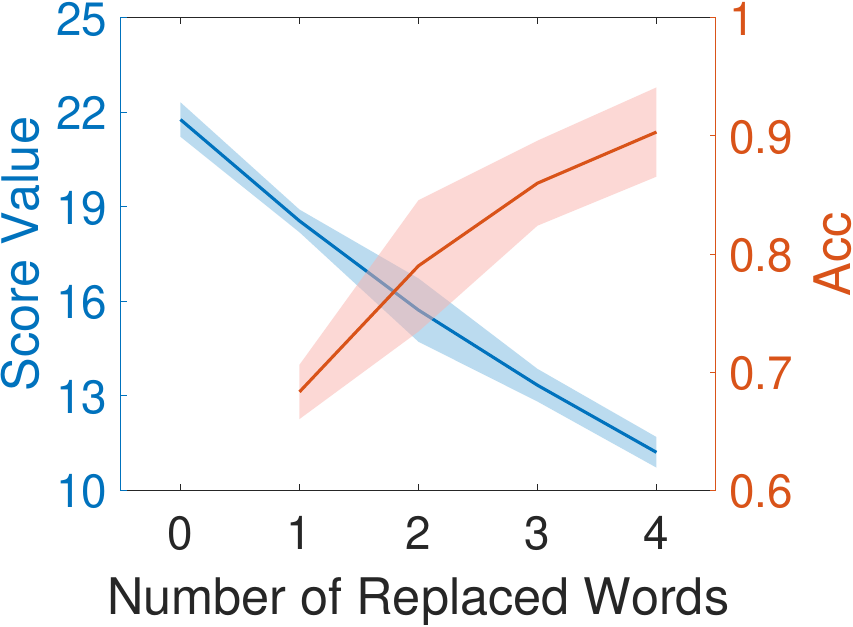}}
    \caption{(a)-(c): The trend of the proposed LEICA score (Score Value) and the corresponding values of Kendall' Tau (K-Tau) when adding different noises with different disturbance degrees to the images. (d): The trend of the LEICA score and Accuracy (Acc) of its judgments \wrt~semantic alignment when gradually replacing the keywords in texts.} 
    \label{exp:fidelity_alignment}
\end{figure}

\subsection{Results on Manually Synthesized Data}\label{sec:toy}

We first manually introduce artifacts into the real-world image-text pairs from the MSCOCO dataset, so that we have oracle judgments for those data that can be used to assess the performance of LEICA.
Specifically, for the effectiveness of LEICA on perceptual image quality, we add noises with multiple degrees into the images while keeping the paired texts unchanged.
We consider several types of noises, including {Gaussian Noise} (GN), {Gaussian Blur} (GB), and {Salt \& Pepper Noise} (SPN). 
\qi{For more complicated distortions, we combine a ``funny mirror'' distortion~\cite{funmirror} with these three noises, named GN+, GB+, and SPN+. Besides, we reconstruct images using the dVAE model of DALL-E~\cite{ramesh2021zero} to obtain a type of model-based distortion.}
\qi{For semantic text-image alignment, we gradually replace the keywords (\ie, those with strong semantic meanings) in the text with random words, or directly use mismatched text for the image, where the paired image stays unaltered.}
To ensure the reliability of our results, we run each experiment five times and calculate their mean values and standard deviations.

\paragraph{Perceptual Quality}
As shown in Figure~\ref{exp:fidelity_alignment}(a)-(c), as the disturbance degree gradually increases, the LEICA score of these images decreases accordingly, showing that the proposed LEICA score can reflect the perceptual quality of the given images.
Furthermore, for each text $x$, we take its paired clean image $y_\text{clean}$ as well as the corresponding noised one $y_\text{noise}$ as a triplet, \ie, $(x, y_\text{noise}, y_\text{clean})$. For each triplet, if the LEICA score of $y_\text{clean}$ is higher than that of $y_\text{noise}$, it is concordant with the oracle judgment. Otherwise, the triplet is discordant. 
Based on this, we calculate Kendall's Tau coefficient for the triplets.
From Figure~\ref{exp:fidelity_alignment}(a)-(c), Kendall's Tau coefficient becomes larger as the disturbance becomes heavier, which further shows that LEICA can reliably measure the perceptual quality of the image.

\paragraph{Semantic Alignment}
See the result in Figure~\ref{exp:fidelity_alignment}(d), as the number of replaced words increases, the LEICA scores on the paired images drop clearly, which demonstrates that our proposed metric is sensitive to the semantic alignment between text and image.
Moreover, the accuracy of making a correct judgment, \ie, predicting a larger score for the paired image with clean text than that with noised text, increases steadily from $70\%$ to more than $90\%$. This indicates that LEICA is able to perceive the mild changes in the semantic meaning of the text, and can also correctly assess the alignment degree between the image-text pairs.

\begin{table}[t]
  \centering
  \caption{Ablation studies on the components of LEICA. The first row of results indicates the performance of LEICA. The last row is the performance of averaged log-likelihood. ``+'' denotes combining the ``funny mirror'' distortion with the current distortion. dVAE: distortion using the dVAE model of DALL-E. Best results are highlighted with \textbf{bold}.}
  \resizebox{1.0\linewidth}{!}
  {
    \begin{tabular}{cc|cccc|ccc|c}
    \toprule
 \multirow{2}[1]{*}{$H$}  & \multirow{2}[1]{*}{$S$} & \multirow{2}[1]{*}{Acc} & \multicolumn{7}{c}{Kendall's Tau ($\uparrow$)}    \\
     \cmidrule{4-10} & & & GN & GB & SPN  & GN+ & GB+ & SPN+ & dVAE\\
    \midrule 
          $\surd$ & $\surd$ &   \textbf{1.00}  &   \textbf{0.86} & \textbf{0.66} & \textbf{0.73} & \textbf{0.70} & \textbf{0.60} & \textbf{0.72} & \textbf{0.44}  \\
          & $\surd$  &   0.96    & 0.60 & 0.40 & 0.60 & 0.66 & 0.32 & \textbf{0.72} & 0.20   \\
          $\surd$ &   &     0.86     & 0.73 & \textbf{0.66}  & 0.60  & 0.54 & 0.56  & 0.64 & 0.42 \\
              &   &    0.54    & 0.33 & -0.20  & 0.40 & 0.43 & 0.20 & 0.40 & -0.04 \\
    \bottomrule
    \end{tabular}%
  }
  \label{exp:ablation_study}%
\end{table}%

\paragraph{Performance Analysis} 
We analyze the effect of components of LEICA, \ie, the indicator function $H$ for perceptual significance and the scoring function $S$ for semantic significance.
Notably, when all these components are removed, LEICA is annealed to the log version of Eq.~(\ref{eq:likelihood}).
\qi{We calculate the accuracy of LEICA on discriminating matched or mismatched text-image pairs with a $1:1$ ratio. We set GN \& GN+ with a variance of $0.2$, GB \& GB+ with a standard deviation of $8$, and SPN \& SPN+ with a noise level of $0.4$.}

As shown in Table~\ref{exp:ablation_study}, when discarding the perceptual significance indicator $H$ from LEICA, the results reduce clearly in all the meta-evaluation metrics, \eg, the accuracy drops from $1.00$ to $0.96$ and the Kendall's Tau coefficient drops from $0.86$ to $0.60$ when adding Gaussian noise (GN). 
Note that the performance drops more severely in the settings related to the image disturbances, indicating that $H$ mainly focuses on the perceptual image quality.
\qi{Besides, when removing the patch-level scoring function $S$, the performance of LEICA drops dramatically in terms of accuracy ($1.00$ to $0.86$), demonstrating its importance in assessing the semantic text-image alignment.}
Lastly, we show the performance of a naive likelihood-based metric, \ie, the averaged log-likelihood, in the last row of the table.
\qi{We find that even with such a naive strategy, it can also achieve the positive Kendall's Tau correlations with human judgment in most settings.
Overall, the above results have shown the effectiveness of the basic likelihood-based evaluation paradigm and the proposed credit assignment strategies \wrt perceptual and semantic significance.}

\begin{table}[t]
  \centering
  \caption{Pearson, Spearman and Kendall's Tau correlations between different evaluation metrics and human judgment. The highest correlation is \textbf{bold}.}
  \resizebox{0.95\linewidth}{!}
  {
    \begin{tabular}{lccc}
    \toprule
    Methods &  Pearson ($\uparrow$) &  Spearman ($\uparrow$)  &  K-Tau ($\uparrow$)  \\
    \midrule
    IS~\cite{salimans2016improved} & 0.7175  &  0.4636   & 0.3091 \\
    FID~\cite{heusel2017gans} & 0.5932  & 0.2363  & 0.1636 \\
    KID~\cite{binkowski2018demystifying} & 0.3039 & 0.2181 & 0.2000 \\
    SOA-C~\cite{hinz2020semantic} & 0.8449 & 0.7363  &  0.5636 \\
    SOA-I~\cite{hinz2020semantic} & 0.8301 & 0.7909  & 0.6000 \\
    CLIP-FID~\cite{kynkaanniemi2022role} &  0.6419  & 0.5636  & 0.4545 \\
    \midrule
    LEICA w/o $H$  & 0.8640  &   0.8727   & 0.7454 \\
    LEICA w/o $S$  &  0.8918  &  0.8636  & 0.7091 \\
    LEICA w/o $e^{\psi/\tau}$  & 0.9101  & 0.9091 & 0.7818 \\
    \midrule
    LEICA (base)&  0.8979 &  0.9181 & 0.7818 \\
    LEICA (large)& \textbf{0.9365} &  \textbf{0.9181} & \textbf{0.7818} \\
    \bottomrule
    \end{tabular}%
    }
  \label{tab:comparison_baseline}%
\end{table}%

\begin{figure}[t]
    \centering
    \includegraphics[width=0.95\linewidth]{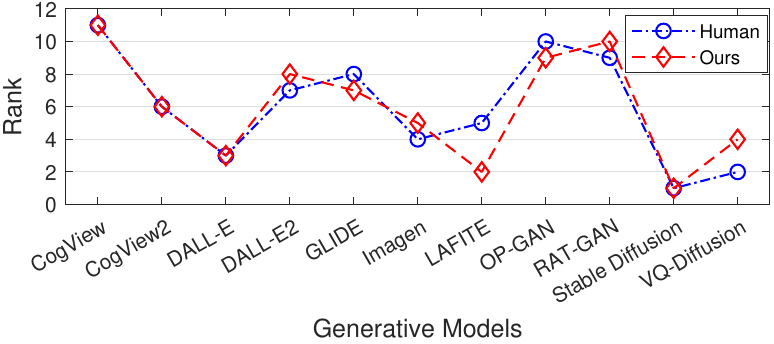}
    \caption{The ranks of each generative model derived by human raters and the proposed evaluation metric. Rank-1 means the best model while rank-11 is the worst one.}
    \label{exp:rank}
\end{figure}

\subsection{Results on Model Generated Data}

We investigate the effectiveness of the proposed metric on generated images. We consider 11 well-known generative models of various types, including Generative Adversarial Networks, Autoregressive models and Diffusion models: CogView~\cite{ding2021cogview}, CogView2~\cite{ding2022cogview2}, DALL-E~\cite{ramesh2021zero}, DALL-E2~\cite{ramesh2022hierarchical}, GLIDE~\cite{nichol2021glide}, Imagen~\cite{saharia2022photorealistic}, LAFITE~\cite{zhou2022towards}, OP-GAN~\cite{hinz2020semantic}, RAT-GAN~\cite{ye2022recurrent}, Stable Diffusion~\cite{rombach2022high}, VQ-Diffusion~\cite{gu2022vector}. Each model synthesizes $5,000$ images, the same as the size of the validation set of MSCOCO.

\begin{figure}[t]
    \centering
    \subfloat[IS]{\includegraphics[width=0.5\linewidth]{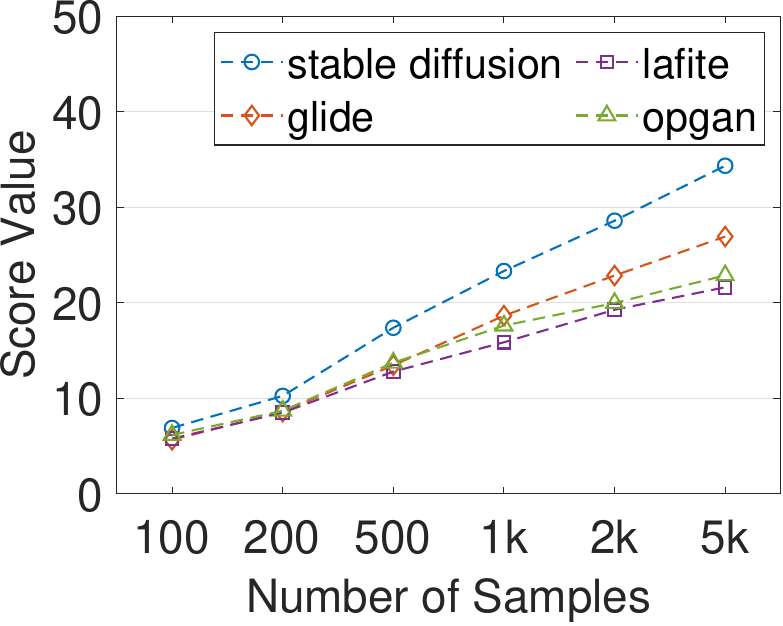}}~~
    \subfloat[FID]{\includegraphics[width=0.5\linewidth]{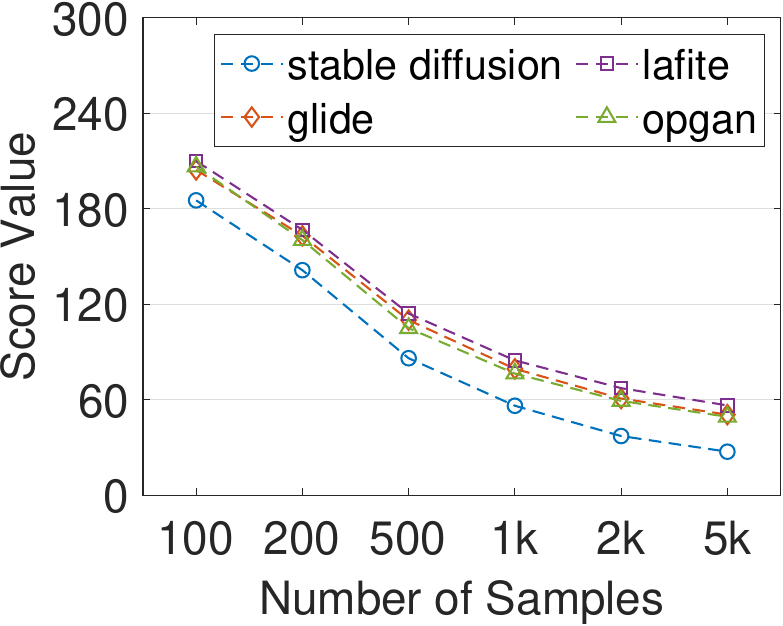}} \\
    \subfloat[SOA-C]{\includegraphics[width=0.5\linewidth]{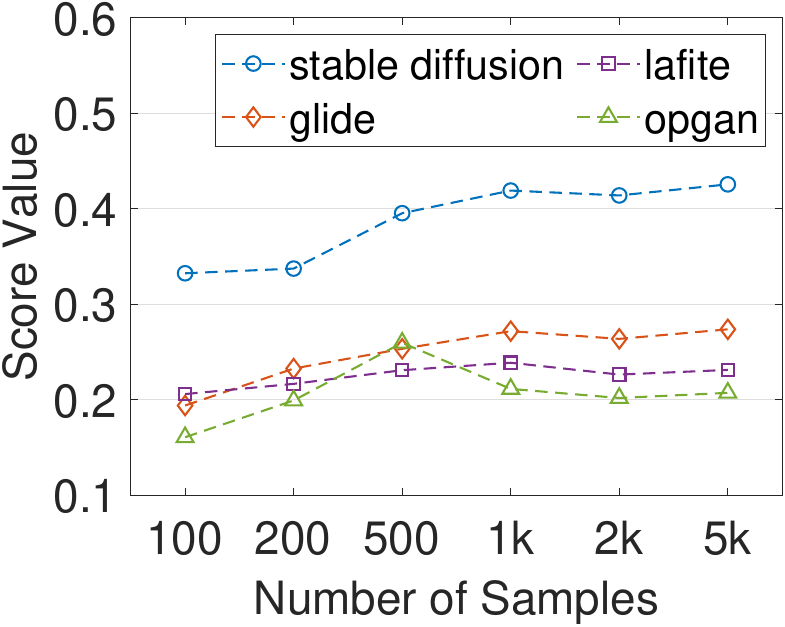}}~~
    \subfloat[Ours]{\includegraphics[width=0.5\linewidth]{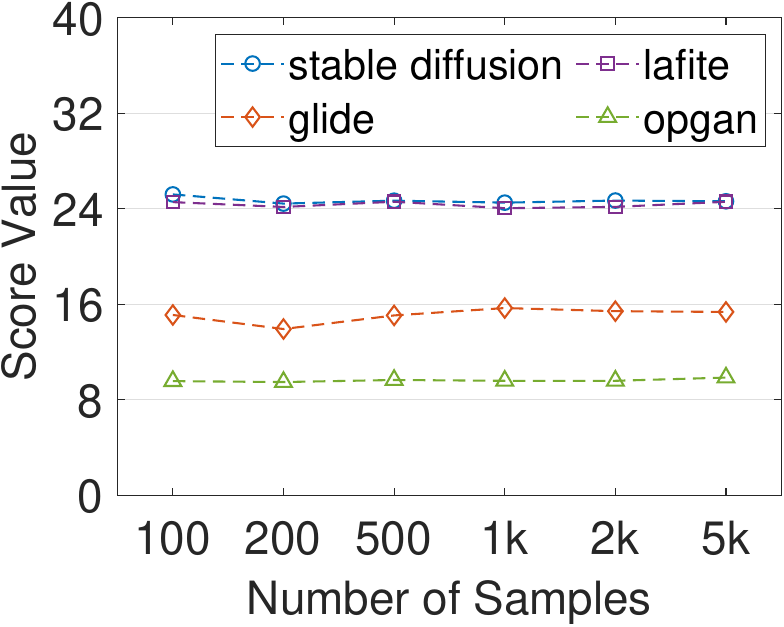}}
    \caption{Impacts of the number of evaluated samples. We test three commonly used evaluation metrics (\ie, IS, FID and SOA-C) as well as the proposed LEICA for the generated images derived from four different generative models.}
    \label{exp:impact_of_num}
\end{figure}

\paragraph{Alignment with Human Judgment}


We conduct a human study to test the alignment between our evaluation metric and human judgment.
Specifically, we randomly choose $100$ texts from the Karpathy's validation split, and obtain corresponding generated images from all $11$ models,
so that each testing sample for our human study contains one text and $11$ images.
\qi{We use a Mean Opinion Score (MOS) evaluation approach to score the test samples, where the score ranges from 1 to 10 in 1-point increments. 
For each testing sample, we ask the raters to score the generated image according to the perceptual image quality and semantic text-image alignment, and emphasize that these two aspects should be equally considered.
Finally, we collect $8,479$ scores for all the $1,100$ testing samples, where each sample has $7.7$ human judgments on average.}

After we obtain the human study results, we conduct a meta-evaluation that compares our metric with six commonly used evaluation metrics in the field of text-to-image generation, \ie, Inception Score (IS) \cite{salimans2016improved}, Fr\'{e}chet Inception Distance (FID) \cite{heusel2017gans}, CLIP-FID~\cite{kynkaanniemi2022role}, Kernel Inception Distance (KID) \cite{binkowski2018demystifying} and Semantic Object Accuracy (SOA) \cite{hinz2020semantic} for class average (SOA-C) and image average (SOA-I). 
For each metric, we compute the Pearson, Spearman, and Kendall's Tau correlation coefficient between the model rankings under its judgment and the human judgment.

As shown in Table~\ref{tab:comparison_baseline}, our evaluation metric consistently outperforms the baseline metrics on all meta-evaluations by a large margin.
Besides, we also investigate the impact of each component in LEICA as in Section \ref{sec:toy}.
From this table, we can draw a similar conclusion as in Table \ref{exp:ablation_study}, where discarding any of these components leads to a decrease in performance.
Nevertheless, all the ablation results of our method are still superior to the compared metrics in terms of the correlation with human judgment.
\qi{Besides, in Table~\ref{tab:comparison_baseline}, we also show the impact of the term $e^{\psi/\tau}$ in Eq.~(\ref{eq:scoring_function}): remove it decreases the performance.}

Moreover, in Figure~\ref{exp:rank}, we plot the model ranking derived from human raters and the ranking based on the proposed LEICA. We observe that the two rankings yield very similar trends, which means our metric is well aligned with human judgment and would not bias to any specific model.
\qi{Lastly, comparing the performance of LEICA (OFA-base) and LEICA (OFA-large), in Table~\ref{tab:comparison_baseline}, we find that a better likelihood estimator will improve the reliability of LEICA, suggesting its scalability to future generative foundation models.}

\begin{table}[t]
  \centering
  \caption{Pearson, Spearman and Kendall's Tau correlations between different evaluation metrics and human judgment on CUB \& Oxford-Flower. The highest correlation is \textbf{bold}.}
  \resizebox{0.85\linewidth}{!}
  {
    \begin{tabular}{lccc}
    \toprule
    Methods &  Pearson ($\uparrow$) &  Spearman ($\uparrow$)  &  K-Tau ($\uparrow$)  \\
    \midrule
     \multicolumn{4}{c}{CUB} \\
    \midrule
    IS~\cite{salimans2016improved} & 0.2636  & 0.3   & 0.2  \\
    FID~\cite{heusel2017gans} & 0.7033  &  \textbf{0.7}  & \textbf{0.6}  \\
    KID~\cite{binkowski2018demystifying} & 0.6806 & 0.6 &  0.4   \\
    LEICA (Ours) &  \textbf{0.8189}  &  \textbf{0.7}  & \textbf{0.6}  \\
    \midrule
         \multicolumn{4}{c}{Oxford-Flower} \\
    \midrule
    IS~\cite{salimans2016improved} & 0.3166    & 0.2   & 0.2  \\
    FID~\cite{heusel2017gans} & 0.4719  &  0.5  & \textbf{0.4}  \\
    KID~\cite{binkowski2018demystifying} & -0.3113 & -0.1 &  0.0   \\
    LEICA (Ours) &  \textbf{0.4880}  &  \textbf{0.6}  & \textbf{0.4}  \\
    \bottomrule
    \end{tabular}%
    }
  \label{tab:cub_flower}%
\end{table}%

\paragraph{Impacts of Number of Evaluated Samples}
In this part, we investigate how the number of evaluation samples influences the evaluation results for our LEICA metric and three baseline metrics, including IS, FID and SOA-C.
We calculate the scores of these metrics using different numbers of generated samples ranging from $100$ to $5,000$ for four different generative models.
In Figure~\ref{exp:impact_of_num}, the values of IS and FID are severely affected by the number of samples. SOA-C performs more stable than IS and FID, but still suffers from fluctuation when the number of samples is changed. 
Therefore, these metrics have to use a very large number of samples to make their evaluation results stable.
In contrast, the scores from the proposed LEICA are very stable, which means that our metric is able to evaluate the performance of the generative models using only a few (one hundred) generated samples, making it very efficient in practice.

\qi{\paragraph{Evaluating Samples from the Same Model \& Text}
Besides accessing different generative models, we further investigate LEICA in another useful scenario, \ie, assessing different samples generated from the same model using the same text.
Specifically, 
we randomly choose $20$ texts from COCO and generate $5$ images for each text with Stable Diffusion (SD) and DALL-E (DE), respectively. Two raters are asked to score these images, and the Pearson correlations between their judgments are $0.203$ (SD) and $0.298$ (DE), showing that this is a very challenging setting even for humans.
Nevertheless, LEICA still achieves positive correlations with these raters (Pearson: $0.105$ for SD and $0.257$ for DE), indicating that our proposed metric has the ability to evaluate individual samples in this setting.
}

\begin{figure}[t]
    \centering
    \includegraphics[width=0.85\linewidth]{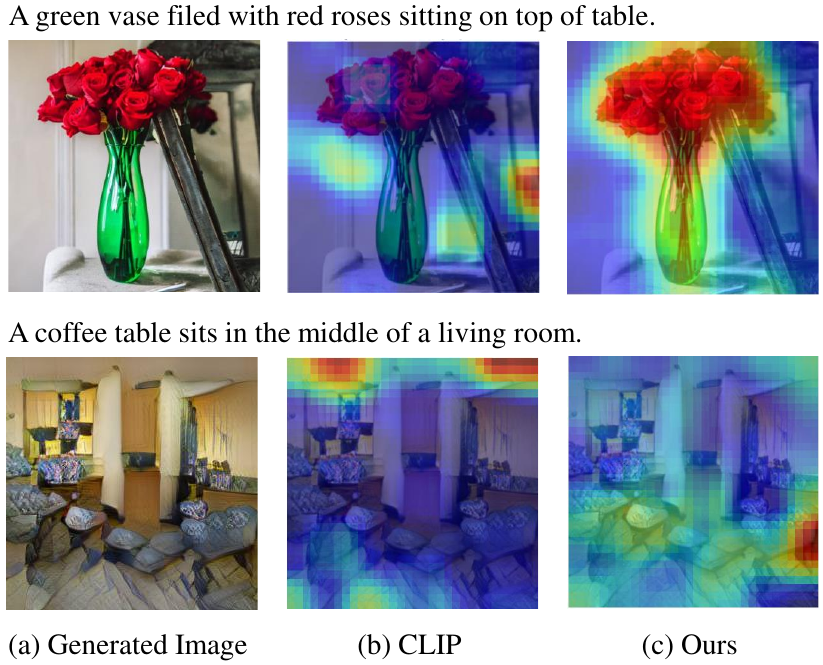}
    \caption{The semantic significance maps obtained from the original CLIP and our modified CLIP on generated images.}
    \label{fig:patch_mask}
\end{figure}

\qi{\paragraph{Evaluation Performance on Specific Domains}
In addition to COCO, in Table~\ref{tab:cub_flower}, we also provide results on the CUB dataset, where LEICA achieves a 0.819 Pearson correlation with the human judgment, clearly better than IS (0.264), FID (0.703) and KID (0.681). SOA is not compared as it is designed only for COCO.
We further evaluate LEICA on Oxford-Flower using the same setting, and it still yields the best performance (\eg, Pearson: 0.488), while IS, FID and KID achieve 0.317, 0.472, and -0.311, respectively.
This demonstrates the generalization ability of LEICA, especially in class-specific domains.
}

\paragraph{Visualization Results}
\qi{We visualize the patch-level alignment performance of the original CLIP and our modified model.}
Figure~\ref{fig:patch_mask} shows that our modified CLIP is able to capture the patches semantically aligned with the given texts in a well-synthesized image (row 1), while suppressing all the regions for a low-quality image (row 2).
\qi{However, the original CLIP fails to capture the semantically significant regions of images described in the given texts.}
These results verify the effectiveness of our modified CLIP.



\section{Conclusion}

In this paper, we propose a likelihood-based metric for text-to-image evaluation based on the intuition that
a good likelihood estimator will reflect how likely a generated image is to exist in the real world given the text.
Besides, we propose two credit assignment strategies \wrt perceptual significance and semantic significance.
The former eliminates the effect of the perceptually-insignificant image regions, 
which may not affect the overall perceptual quality of a generated image but severely drag down its overall likelihood.
The latter down-weights the effect of the background regions with low semantic significance, 
which are usually not related to the text but account for the largest portion of the image.
We evaluate our metric on \textbf{11} models and \textbf{3} datasets, showing that it is effective, efficient, and stable.

{\small
\bibliographystyle{ieee_fullname}
\bibliography{egbib}

\begin{thebibliography}{10}\itemsep=-1pt

\bibitem{funmirror}
\url{https://github.com/kaustubh-sadekar/FunMirrors}.

\bibitem{binkowski2018demystifying}
Miko{\l}aj Bi{\'n}kowski, Danica~J Sutherland, Michael Arbel, and Arthur
  Gretton.
\newblock Demystifying mmd gans.
\newblock {\em ICLR}, 2018.

\bibitem{chen2020simple}
Ting Chen, Simon Kornblith, Mohammad Norouzi, and Geoffrey Hinton.
\newblock A simple framework for contrastive learning of visual
  representations.
\newblock In {\em ICML}, pages 1597--1607, 2020.

\bibitem{deng2009imagenet}
Jia Deng, Wei Dong, Richard Socher, Li-Jia Li, Kai Li, and Li Fei-Fei.
\newblock Imagenet: A large-scale hierarchical image database.
\newblock In {\em CVPR}, pages 248--255, 2009.

\bibitem{ding2021cogview}
Ming Ding, Zhuoyi Yang, Wenyi Hong, Wendi Zheng, Chang Zhou, Da Yin, Junyang
  Lin, Xu Zou, Zhou Shao, Hongxia Yang, et~al.
\newblock Cogview: Mastering text-to-image generation via transformers.
\newblock {\em NeurIPS}, pages 19822--19835, 2021.

\bibitem{ding2022cogview2}
Ming Ding, Wendi Zheng, Wenyi Hong, and Jie Tang.
\newblock Cogview2: Faster and better text-to-image generation via hierarchical
  transformers.
\newblock {\em NeurIPS}, 2022.

\bibitem{esser2021taming}
Patrick Esser, Robin Rombach, and Bjorn Ommer.
\newblock Taming transformers for high-resolution image synthesis.
\newblock In {\em CVPR}, pages 12873--12883, 2021.

\bibitem{gu2022vector}
Shuyang Gu, Dong Chen, Jianmin Bao, Fang Wen, Bo Zhang, Dongdong Chen, Lu Yuan,
  and Baining Guo.
\newblock Vector quantized diffusion model for text-to-image synthesis.
\newblock In {\em CVPR}, pages 10696--10706, 2022.

\bibitem{heusel2017gans}
Martin Heusel, Hubert Ramsauer, Thomas Unterthiner, Bernhard Nessler, and Sepp
  Hochreiter.
\newblock Gans trained by a two time-scale update rule converge to a local nash
  equilibrium.
\newblock {\em NeurIPS}, 2017.

\bibitem{hinz2020semantic}
Tobias Hinz, Stefan Heinrich, and Stefan Wermter.
\newblock Semantic object accuracy for generative text-to-image synthesis.
\newblock {\em IEEE TPAMI}, 2020.

\bibitem{kynkaanniemi2022role}
Tuomas Kynk{\"a}{\"a}nniemi, Tero Karras, Miika Aittala, Timo Aila, and Jaakko
  Lehtinen.
\newblock The role of imagenet classes in fr$\backslash$'echet inception
  distance.
\newblock {\em ICLR}, 2023.

\bibitem{lin2014microsoft}
Tsung-Yi Lin, Michael Maire, Serge Belongie, James Hays, Pietro Perona, Deva
  Ramanan, Piotr Doll{\'a}r, and C~Lawrence Zitnick.
\newblock Microsoft coco: Common objects in context.
\newblock In {\em ECCV}, 2014.

\bibitem{nichol2021glide}
Alex Nichol, Prafulla Dhariwal, Aditya Ramesh, Pranav Shyam, Pamela Mishkin,
  Bob McGrew, Ilya Sutskever, and Mark Chen.
\newblock Glide: Towards photorealistic image generation and editing with
  text-guided diffusion models.
\newblock {\em ICML}, 2022.

\bibitem{nilsback2008automated}
Maria-Elena Nilsback and Andrew Zisserman.
\newblock Automated flower classification over a large number of classes.
\newblock In {\em 2008 Sixth Indian Conference on Computer Vision, Graphics \&
  Image Processing}, pages 722--729. IEEE, 2008.

\bibitem{radford2021learning}
Alec Radford, Jong~Wook Kim, Chris Hallacy, Aditya Ramesh, Gabriel Goh,
  Sandhini Agarwal, Girish Sastry, Amanda Askell, Pamela Mishkin, Jack Clark,
  et~al.
\newblock Learning transferable visual models from natural language
  supervision.
\newblock In {\em ICML}, pages 8748--8763, 2021.

\bibitem{ramesh2022hierarchical}
Aditya Ramesh, Prafulla Dhariwal, Alex Nichol, Casey Chu, and Mark Chen.
\newblock Hierarchical text-conditional image generation with clip latents.
\newblock {\em arXiv preprint arXiv:2204.06125}, 2022.

\bibitem{ramesh2021zero}
Aditya Ramesh, Mikhail Pavlov, Gabriel Goh, Scott Gray, Chelsea Voss, Alec
  Radford, Mark Chen, and Ilya Sutskever.
\newblock Zero-shot text-to-image generation.
\newblock In {\em ICML}, pages 8821--8831, 2021.

\bibitem{ravuri2019classification}
Suman Ravuri and Oriol Vinyals.
\newblock Classification accuracy score for conditional generative models.
\newblock {\em NeurIPS}, 32, 2019.

\bibitem{redmon2018yolov3}
Joseph Redmon and Ali Farhadi.
\newblock Yolov3: An incremental improvement.
\newblock {\em arXiv preprint arXiv:1804.02767}, 2018.

\bibitem{ren2019likelihood}
Jie Ren, Peter~J Liu, Emily Fertig, Jasper Snoek, Ryan Poplin, Mark Depristo,
  Joshua Dillon, and Balaji Lakshminarayanan.
\newblock Likelihood ratios for out-of-distribution detection.
\newblock {\em Advances in neural information processing systems}, 32, 2019.

\bibitem{rombach2022high}
Robin Rombach, Andreas Blattmann, Dominik Lorenz, Patrick Esser, and Bj{\"o}rn
  Ommer.
\newblock High-resolution image synthesis with latent diffusion models.
\newblock In {\em CVPR}, pages 10684--10695, 2022.

\bibitem{saharia2022photorealistic}
Chitwan Saharia, William Chan, Saurabh Saxena, Lala Li, Jay Whang, Emily
  Denton, Seyed Kamyar~Seyed Ghasemipour, Burcu~Karagol Ayan, S~Sara Mahdavi,
  Rapha~Gontijo Lopes, et~al.
\newblock Photorealistic text-to-image diffusion models with deep language
  understanding.
\newblock {\em NeurIPS}, 2022.

\bibitem{salimans2016improved}
Tim Salimans, Ian Goodfellow, Wojciech Zaremba, Vicki Cheung, Alec Radford, and
  Xi Chen.
\newblock Improved techniques for training gans.
\newblock {\em NeurIPS}, 2016.

\bibitem{szegedy2016rethinking}
Christian Szegedy, Vincent Vanhoucke, Sergey Ioffe, Jon Shlens, and Zbigniew
  Wojna.
\newblock Rethinking the inception architecture for computer vision.
\newblock In {\em CVPR}, pages 2818--2826, 2016.

\bibitem{theis2015note}
Lucas Theis, A{\"a}ron van~den Oord, and Matthias Bethge.
\newblock A note on the evaluation of generative models.
\newblock {\em ICLR}, 2016.

\bibitem{van2016pixel}
A{\"a}ron Van Den~Oord, Nal Kalchbrenner, and Koray Kavukcuoglu.
\newblock Pixel recurrent neural networks.
\newblock In {\em ICML}, pages 1747--1756, 2016.

\bibitem{van2017neural}
Aaron Van Den~Oord, Oriol Vinyals, et~al.
\newblock Neural discrete representation learning.
\newblock {\em NeurIPS}, 2017.

\bibitem{vaswani2017attention}
Ashish Vaswani, Noam Shazeer, Niki Parmar, Jakob Uszkoreit, Llion Jones,
  Aidan~N Gomez, {\L}ukasz Kaiser, and Illia Polosukhin.
\newblock Attention is all you need.
\newblock {\em NeurIPS}, 30, 2017.

\bibitem{wah2011caltech}
Catherine Wah, Steve Branson, Peter Welinder, Pietro Perona, and Serge
  Belongie.
\newblock The caltech-ucsd birds-200-2011 dataset.
\newblock 2011.

\bibitem{wang2022ofa}
Peng Wang, An Yang, Rui Men, Junyang Lin, Shuai Bai, Zhikang Li, Jianxin Ma,
  Chang Zhou, Jingren Zhou, and Hongxia Yang.
\newblock Ofa: Unifying architectures, tasks, and modalities through a simple
  sequence-to-sequence learning framework.
\newblock In {\em ICML}, pages 23318--23340, 2022.

\bibitem{xu2018attngan}
Tao Xu, Pengchuan Zhang, Qiuyuan Huang, Han Zhang, Zhe Gan, Xiaolei Huang, and
  Xiaodong He.
\newblock Attngan: Fine-grained text to image generation with attentional
  generative adversarial networks.
\newblock In {\em CVPR}, pages 1316--1324, 2018.

\bibitem{ye2022recurrent}
Senmao Ye, Fei Liu, and Minkui Tan.
\newblock Recurrent affine transformation for text-to-image synthesis.
\newblock {\em arXiv preprint arXiv:2204.10482}, 2022.

\bibitem{yu2022scaling}
Jiahui Yu, Yuanzhong Xu, Jing~Yu Koh, Thang Luong, Gunjan Baid, Zirui Wang,
  Vijay Vasudevan, Alexander Ku, Yinfei Yang, Burcu~Karagol Ayan, et~al.
\newblock Scaling autoregressive models for content-rich text-to-image
  generation.
\newblock {\em arXiv preprint arXiv:2206.10789}, 2022.

\bibitem{yuan2021bartscore}
Weizhe Yuan, Graham Neubig, and Pengfei Liu.
\newblock Bartscore: Evaluating generated text as text generation.
\newblock {\em NeurIPS}, pages 27263--27277, 2021.

\bibitem{zhang2021cross}
Han Zhang, Jing~Yu Koh, Jason Baldridge, Honglak Lee, and Yinfei Yang.
\newblock Cross-modal contrastive learning for text-to-image generation.
\newblock In {\em CVPR}, pages 833--842, 2021.

\bibitem{zhang2017stackgan}
Han Zhang, Tao Xu, Hongsheng Li, Shaoting Zhang, Xiaogang Wang, Xiaolei Huang,
  and Dimitris~N Metaxas.
\newblock Stackgan: Text to photo-realistic image synthesis with stacked
  generative adversarial networks.
\newblock In {\em ICCV}, pages 5907--5915, 2017.

\bibitem{zhou2022towards}
Yufan Zhou, Ruiyi Zhang, Changyou Chen, Chunyuan Li, Chris Tensmeyer, Tong Yu,
  Jiuxiang Gu, Jinhui Xu, and Tong Sun.
\newblock Towards language-free training for text-to-image generation.
\newblock In {\em CVPR}, pages 17907--17917, 2022.

\bibitem{zhu2019dm}
Minfeng Zhu, Pingbo Pan, Wei Chen, and Yi Yang.
\newblock Dm-gan: Dynamic memory generative adversarial networks for
  text-to-image synthesis.
\newblock In {\em CVPR}, pages 5802--5810, 2019.

\end{thebibliography}
}

\end{document}